\pdfoutput=1
\documentclass[11pt]{article}

\usepackage[final]{acl}

\usepackage{times}
\usepackage{latexsym}
\usepackage{graphicx}
\usepackage{epstopdf}
\usepackage{stfloats}
\usepackage{booktabs}
\usepackage{bm}
\usepackage{paralist}
\usepackage{tabularx}
\usepackage{amsthm,amsmath,amssymb}
 
\usepackage{mathrsfs}

\usepackage{colortbl}  %彩色表格需要加载的宏包
\usepackage{xcolor}
\usepackage{array}   %对表列和表格线的设置需要用到array宏包

\usepackage{makecell}
\usepackage{pifont}       % \ding{xx}
\usepackage{bbding}       % 
\usepackage{fontawesome}  % \faCheck,\faTimes
\usepackage{multirow}
\usepackage{longtable}
\usepackage{listings}

\lstdefinestyle{prompt}{
  basicstyle=\ttfamily\small,
  breaklines=true,
  frame=tb,
  columns=fullflexible,
  backgroundcolor=\color[gray]{0.95},
  escapeinside={(*}{*)}
}

\newcommand{\okmark}{{\textbf{\textcolor[rgb]{0.1, 0.5, 0.1}{$\checkmark$}}}}
\newcommand{\ngmark}{{\textbf{\color{red}{\ding{55}}}}}
\usepackage[T1]{fontenc}
\usepackage[utf8]{inputenc}

\usepackage{microtype}

\usepackage{inconsolata}

\title{Dynamic Planning for LLM-based Graphical User Interface Automation}

\author{Shaoqing Zhang$^{1}$,
  Zhuosheng Zhang$^{2}$, Kehai Chen$^{1}$\thanks{ Corresponding authors}, Xinbei Ma$^{2}$, \\\textbf{Muyun Yang}$^{1}$, \textbf{Tiejun Zhao}$^{1}$, \textbf{Min Zhang}$^{1}$\\
  $^1$School of Computer Science and Technology, Harbin Institute of Technology, China,\\ 
  $^2$School of Electronic Information and Electrical Engineering, Shanghai Jiao Tong University, China \\
  \texttt{\{23S003064\}@stu.hit.edu.cn} \\
  \texttt{\{zhangzs, sjtumaxb\}@sjtu.edu.cn} \\
  \texttt{\{chenkehai, yangmuyun, tjzhao, zhangmin2021\}@hit.edu.cn}
  }

\begin{document}
\maketitle
\begin{abstract}
The advent of large language models (LLMs) has spurred considerable interest in advancing autonomous LLMs-based agents, particularly in intriguing applications within smartphone graphical user interfaces (GUIs).
When presented with a task goal, these agents typically emulate human actions within a GUI environment until the task is completed. 
However, a key challenge lies in devising effective plans to guide action prediction in GUI tasks, though planning have been widely recognized as effective for decomposing complex tasks into a series of steps. 
Specifically, given the dynamic nature of environmental GUIs following action execution, it is crucial to dynamically adapt plans based on environmental feedback and action history.
We show that the widely-used ReAct approach fails due to the excessively long historical dialogues. 
To address this challenge, we propose a novel approach called Dynamic Planning of Thoughts (D-PoT) for LLM-based GUI agents.
D-PoT involves the dynamic adjustment of planning based on the environmental feedback and execution history.
Experimental results reveal that the proposed D-PoT significantly surpassed the strong GPT-4V baseline by +12.7\% (34.66\% $\rightarrow$ 47.36\%) in accuracy. 
The analysis highlights the generality of dynamic planning in different backbone LLMs, as well as the benefits in mitigating hallucinations and adapting to unseen tasks. 
Code is available at \url{https://github.com/sqzhang-lazy/D-PoT}.
%confines itself to static methodologies. LLMs-based agents either act without a plan or formulate a plan at the beginning of a task and then adhere to it statically. This approach may lead LLMs-based agents to predict erroneous actions when faced with unexpected situations.
% However, current studies often confine themselves to static plans or lack specific plans entirely. the LLMs-based agent either acts without a plan or formulates a plan at the beginning of a task and then adheres to it statically, which can lead LLMs-based agent to take the wrong action when they encounter an unexpected situation.
% However, it remains an open challenge how to generate effective plans to guide the execution prediction.
% However, current studies often confine themselves to static plans or lack specific plans entirely, which can lead LLMs-based agent to take the wrong action when they encounter an unexpected situation. 
%Given that the environment evolves following action execution, the imperative is to adapt plans dynamically based on environmental feedback and action history. 
%Therefore, we propose Dynamic Planning of Thoughts (D-PoT), a novel approach designed to cultivate dynamic planning. 
%D-PoT involves the dynamic adjustment of planning based on feedback from the environment and execution history.

\end{abstract}

\section{Introduction}

Building autonomous agents capable of assisting humans in addressing real-world challenges has long been a central pursuit of artificial intelligence research~\citep{Searle_1972, Wooldridge_Jennings_1995, Maes_1994}. 
Recently, there has been a surge in exploration within the realm of autonomous agents, fueled largely by the emergence of large language models (LLMs)~\citep{Chowdhery, Wei_Wang, zhong2024understanding}. %, achiam2023gpt}. 
% These LLM-based agents have shown promising opportunities to output instructions in natural language to simulate operating a smartphone via perceiving distinct environments, formulating plans. 
% A prevalent scenario is smartphone graphical user interface (GUI) automation, where agents are tasked with controlling smartphones to complete complex instructions through multi-turn interactions~\citep{rawles2023android, wen2023empowering}. 
One prevalent application scenario involves automating graphical user interfaces (GUIs) on smartphones, where LLMs are tasked with perceiving smartphone GUIs and sequentially predicting action commands until the task is completed~\citep{rawles2023android, zhang2023appagent}.
% Consequently, the adoption of LLM-based methods has significantly propelled the capabilities of such agents ~\cite{yan2023gpt4v,wen2023empowering}.
% As a result, the LLMs-based method has greatly advanced the agent~\cite{yan2023gpt4v,wen2023empowering}.

\begin{figure}[t]
    \centering
    \includegraphics[scale=0.365]{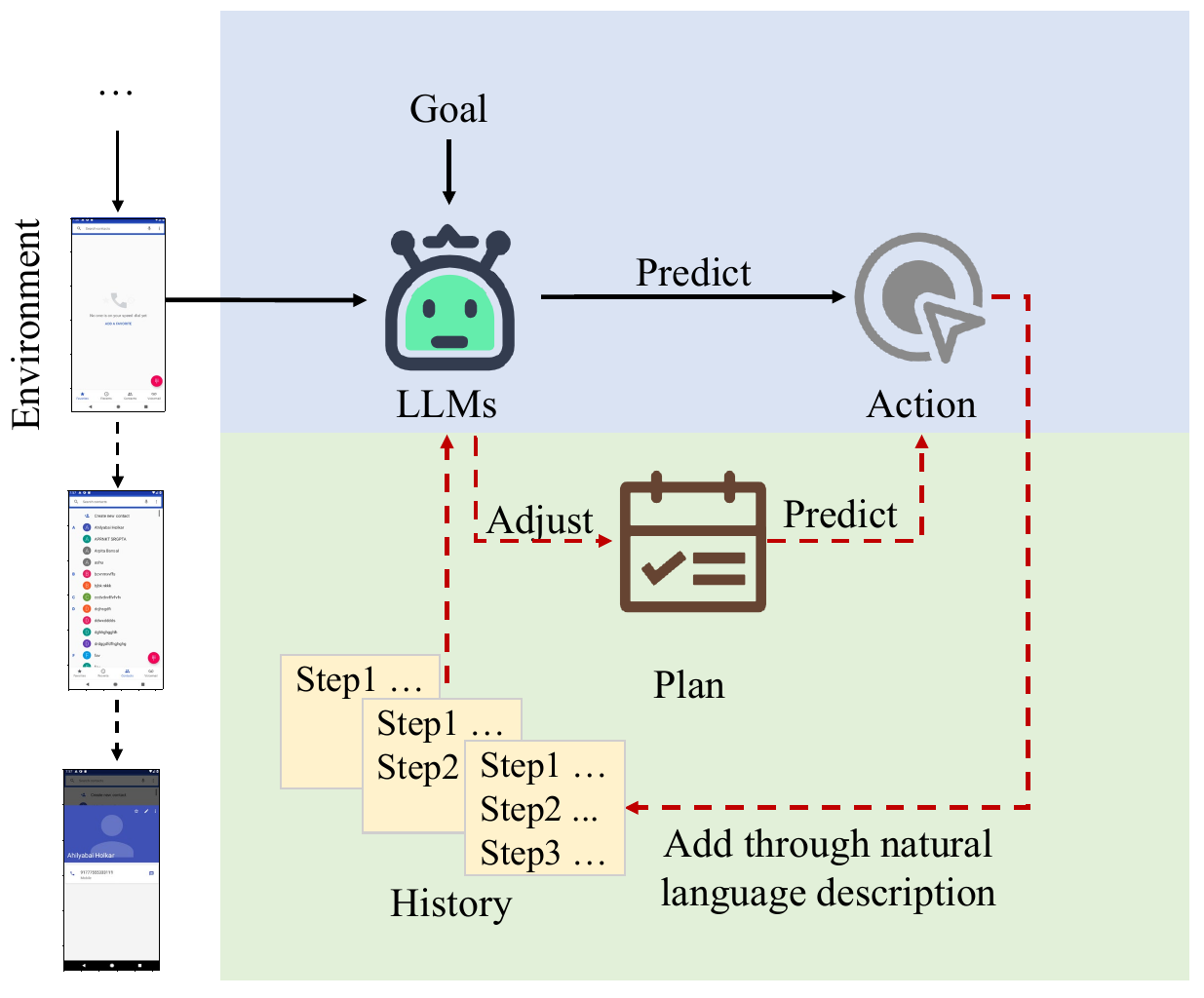}
    % \vspace{-2mm}
    \caption{The proposed dynamic planning method incorporates the execution history to adjust the plan to predict the action and subsequently supplements the execution history with the predicted action.}
    \label{fig:static_and_dynamic}
    \vspace{-2mm}
\end{figure}

While previous studies have made significant strides by enhancing environment perception through fine-grained GUI grounding~\citep{zhang2023look, hong2023cogagent,yan2023gpt4v,zhang2023appagent,cheng2024seeclick,you2024ferret,liu2024memlong}, there has been limited focus on the planning capabilities of GUI agents.
Evidence suggests that decomposing a complex task into a series of plans is effective in eliciting the ability of LLMs within agent systems \citep{zhu2024knowagent,huang2024understanding,song2023llm, gui2024vector}. Additionally, given that the environment evolves based on action predictions, it is crucial to dynamically adapt plans based on environmental feedback and execution history. 

Existing LLMs-based GUI agents typically takes actions directly prior planning or adjustment of plans based on environmental feedback (e.g., new GUI screenshot) and execution history (e.g., previous steps described in natural language). 
For instance, as depicted in Figure~\ref{fig:static_and_dynamic}, the static\footnote{The methods are static due to be not aware of historical information during task execution.} methods (black data stream) directly predicts actions based on the screenshot and goal. Those approaches struggles to handle complex real-world scenarios, where users often adjust subsequent actions based on past steps. We will show that the widely-used ReAct approach \citep{yao2023react} fails due to excessively long historical dialogues, revealing its inadequacy in handling complex real-world scenarios (Section \ref{sec:challenge}). 

To address the challenge above, we propose a novel method called Dynamic Planning of Thoughts (D-PoT) method to enable the LLM-based agent to formulate effective plans based on environment feedback and execution history during task execution (with dashed lines in Figure~\ref{fig:static_and_dynamic}).
Concretely, D-PoT dynamically adjusts its plans by incorporating new screenshots and execution history throughout the goal attainment process. Moreover, our proposed method allows for continuous refinement of the current plan, ensuring persistent optimization until the desired goal is achieved. 
Experimental results demonstrate that our planning mechanism substantially improves the task performance.
Additionally, analysis highlights the efficacy of dynamic planning in mitigating hallucinations and adapting to unseen tasks. 

Our key contributions are as follows:

(i) D-PoT dynamically formulates plans and selects steps for action prediction based on the new screenshots and execution history, thereby advancing the LLMs-based agent.

(ii) D-PoT achieves a notable improvement in accuracy scores of +12.7\% (34.66\% $\rightarrow$ 47.36\%) compared with the strong GPT-4V baseline.

(iii) Analysis highlights the efficacy of dynamic planning in not only enhancing action prediction accuracy but also in in mitigating hallucinations and adapting to previously unfamiliar tasks.

\section{Related Work}

Our work is related to LLMs-based GUI agents. This section will first review the recent progress of the work on building the GUI agents and then discuss the planning mechanism of the agents.

\subsection{LLMs-based GUI Agent}
% LLMs have spurred considerable interest in the realm of language agents, which adeptly adhere to language instructions and execute actions in interactive environments. 

LLMs have spurred considerable interest in the realm of language agents. 
% Generally, LLMs possess robust zero-shot and few-shot adaptability, enabling them to comprehend input within specific scenarios and inference within prescribed output formats. 
Notable examples include AutoGPT~\citep{yang2023auto}, HuggingGPT~\citep{shen2023hugginggpt}, and MetaGPT~\citep{Group}, all of which explored the integration of LLMs as the core of agents~\citep{ji2024feature}.

This work focuses on ultilizing LLMs as intelligent assistants for smartphones. 
% The task will define the output format for inference in advance, typically mirroring the operational instructions for smartphones. It necessitates LLMs to anticipate the output action by comprehending the current screen input and the task goal. 
These assistants are crafted to assist people in accomplishing their daily tasks and meeting life's requirements, especially enhancing accessibility for individuals with disabilities.
% These assistants are designed to help humans in fulfilling the daily tasks and requirements of life, which are especially beneficial for people with disabilities. 
Notably, the advent of multi-modal LLMs such as GPT-4V, showcasing robust image understanding capabilities~\citep{yang2023dawn}, has prompted previous research to predominantly concentrate on comprehending GUI interactions. For instance, MM-Navigator delved into leveraging optical character recognition (OCR) parsing to enhance GPT-4V's GUI comprehension~\citep{yan2023gpt4v}, while AppAgent reinforced the understanding of Application GUI elements by introducing the roles of distinct GUI~\citep{zhang2023appagent}. In addition to these, CogAgent, Auto-GUI and CoCo-Agent fine-tuned the agent's understanding of GUI to enhance performance~\citep{hong2023cogagent, zhang2023you, ma2024coco}. With it comes risk, and these agents have also suffered many attacks~\citep{ma2024caution}.

In contrast to the prior research that concentrates on multimodal perception, our work focuses on the planning mechanism to enhance the LLMs proficiency in planning and effectively tackle multi-step tasks on smartphones.

\subsection{Planning Mechanisms for LLMs}

LLMs have shown considerable potential in constructing agents with strong capabilities in following instructions and maintaining coherent chains of thought (CoT) via solving complex problems~\citep{Wei_Wang, kojima2023large, zhang2022automatic}. Notably, the CoT prompting technique has enabled LLMs to engage in effective step-by-step problem-solving process~\citep{huang2023reasoning, yao2023tree, wang2023selfconsistency, chen2023program}. To address more complex problems, divide-and-conquer prompting strategies have been proposed, e.g., dividing problems into manageable steps~\citep{zhou2023leasttomost, lee2023recursion} or sequential solutions~\citep{wang2023planandsolve}.

The research above mainly focuses on enhancing the reasoning abilities of LLMs. However, the ReAct~\citep{yao2023react} has inspired researchers to explore more suitable ways for LLMs to complete agentic tasks by leveraging their reasoning abilities. This approach involves LLMs first observing and reasoning before taking action, such as utilizing external tools to identify and rectify errors ~\citep{gou2023critic, shinn2023reflexion}, or planning before executing~\citep{wang2023planandsolve, hao2023reasoning}.

% Inspired by the progress above, we are motivated to design a novel dynamic planning mechanism for the GUI tasks. We formulate plans by integrating both execution history and environmental cues to guide its actions. Meanwhile, we devise plans in response to environmental stimuli, with these plans serving not only as assessments of present decisions but also as anticipations of future actions.

\section{Investigating the Necessity of Dynamic Planning in GUI Agent}\label{sec:challenge}
\subsection{Challenge of GUI Automation}
GUI automation is a long-episode task where the LLM first receives a goal and an initial screen. To achieve this goal, it must navigate through multiple screens continuously until the task is complete. This presents a challenge for the LLM, requiring it to understand the current progress of the task and the execution history to avoid performing redundant actions in similar environments.
\subsection{ReAct Fails Due to the Excessively Long Historical Dialogues}
ReAct is a widely used method in the LLM-based agent~\citep{yao2023react}. It encourages the LLM to think before taking action when encountering a new environment. Each round of input includes all previous thoughts and actions. We experiment with using ReAct for the GUI Automation task. We sampled 20 tasks from a general dataset and conducted experiments using GPT-4V. The experimental setup is detailed in Section~\ref{sec:implementation_detail}.

\begin{table}[ht]
\small
    \centering
    \setlength{\tabcolsep}{2mm}
    \begin{tabular}{lrrr}
    \toprule
        \textbf{\makecell{History \\Length}} & \textbf{Accuracy} & \textbf{\makecell{Inference Cost\\(tokens/ep)}} & \textbf{\makecell{Inference \\Speed(s/ep)}} \\ \midrule
        0 & 32.58 & 22265.3 & 55.1 \\
        1 & 42.42 & 40961.1 & 63.3 \\
        2 & 45.45 & 57176.6 & 77.4 \\
        4 & 44.70 & 80300.2 & 91.1 \\
        $\infty$ & 43.94 & 101151.1 & 119.7 \\
        \midrule
        D-PoT & 45.45 & 25418.7 & 50.3 \\
        \bottomrule
    \end{tabular}
    \caption{History Length refers to the number of dialogue rounds inputted into ReAct, where $``\infty''$ indicates that all historical dialogues are included. Accuracy calculation details can be found in Section~\ref{sec:dataset}. Inference cost represents the average tokens used per episode for API call, and inference speed indicates the number of seconds required to complete each task.}
    \label{tab:pre_exam}
    \vspace{-3.6mm}
\end{table}

Based on the Table~\ref{tab:pre_exam}, we observe that accuracy does not always improve with the increased length of input historical dialogues. The best performance is achieved when the history length is 2. This is likely because, in the GUI Automation task, the input length is substantial, with each round containing at least 2000 tokens. Consequently, the performance does not significantly improve with an increase in the length of historical dialogues.

\begin{figure}[ht]
    \centering
    \includegraphics[scale=0.25]{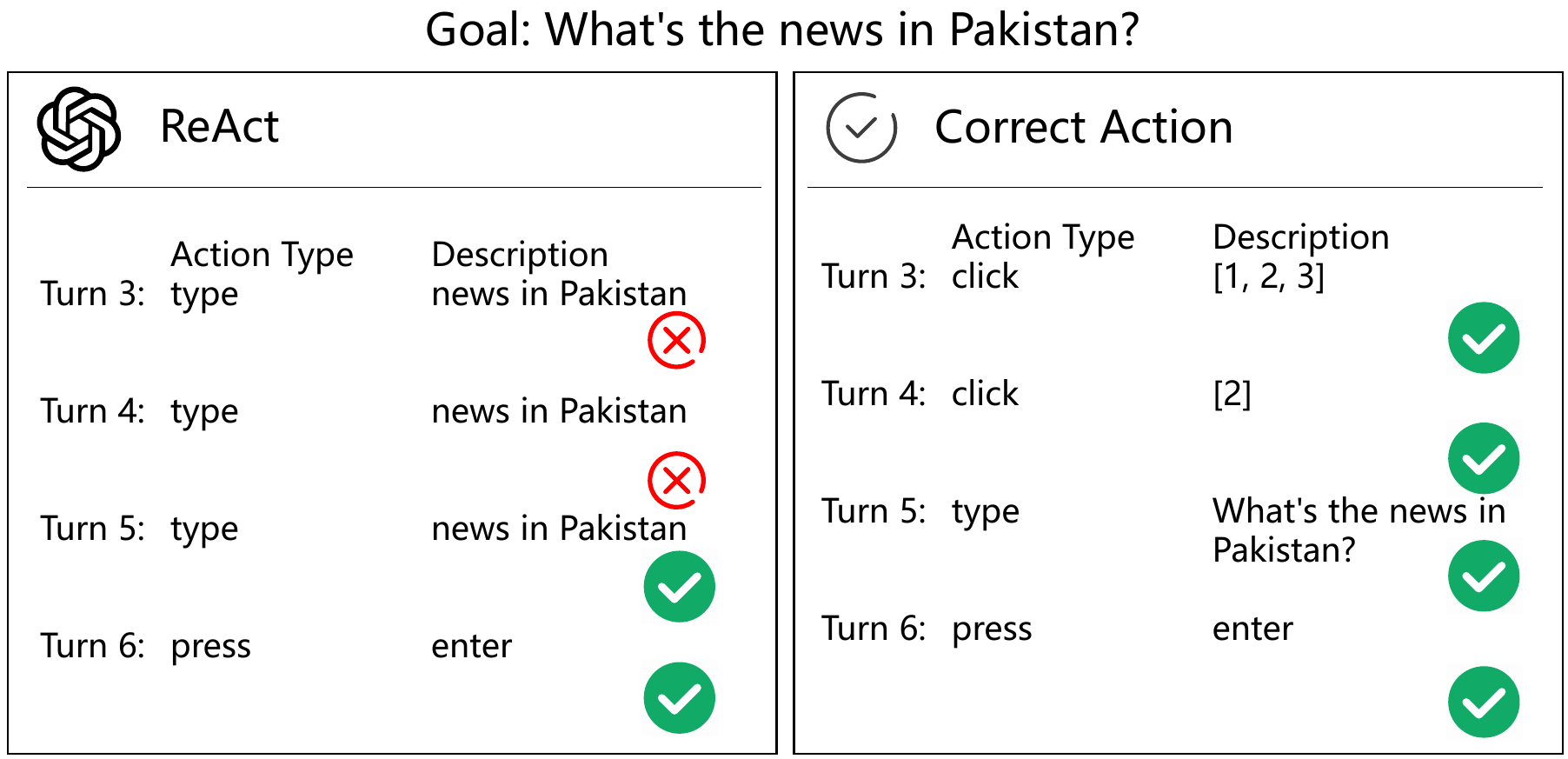}
    \caption{ReAct is misled by incorrect decisions.}
    \label{fig:react}
    \vspace{-3mm}
\end{figure}

\begin{figure*}[ht]
    \centering
    \includegraphics[scale=0.77]{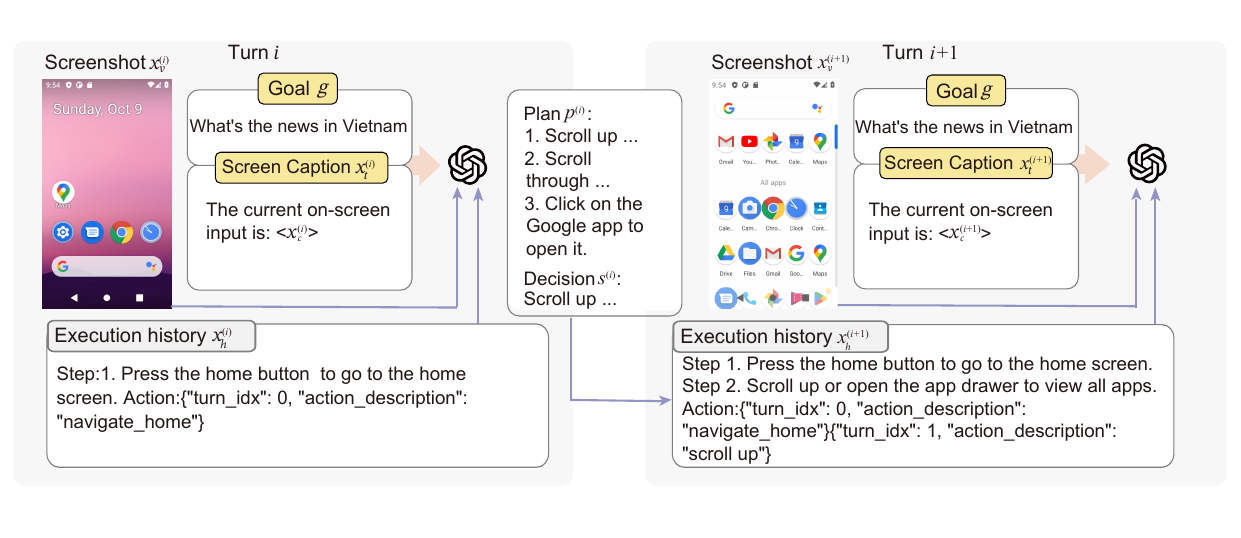}
    \vspace{-8mm}
    \caption{Overview of D-PoT. In turn, $i$, the D-PoT makes a plan based on visual input and textual input, predicts the action to be performed, and then updates the execution history, and then proceeds to the next turn $i+1$.}
    \label{fig:framework}
    \vspace{-3mm}
\end{figure*}

Additionally, ReAct can be misled by incorrect decisions made in the last turn. We speculate that the reason is because ReAct focuses more on the most recent interaction rather than the overall task completion progress. 
For instance, as depicted in Figure~\ref{fig:react}. In the third turn, a "type" action is performed, but it is incorrect. This error persist until the fifth turn when it is finally resolved.

When facing frequent interactions or complex task inputs, reducing inference cost and speeding up inference become critical challenges. The key issue lies in how to effectively input execution history to make dynamic plans and guide LLM-based agents in  understanding task progress.

\section{Method}
In light of the above experimental results and analysis, we propose Dynamic Planning of Thoughts (D-PoT) to mitigate the challenge. On a high level, D-PoT consists of two stages: 

% The proposed D-PoT approach is grounded in dynamic planning based on environment feedback and execution history and includes two stages: 
(1) planning initialization: the LLMs initiate the planning process by generating an overall plan, considering the ultimate goal, current visual input, and prior execution history. The plan helps the LLM grasp the progress of the current task. Once the plan is formulated, the LLMs will select the most plausible step for execution. 
(2) dynamic planning adjustment: the executed step is appended to the execution history. This updated history list then carefully shapes subsequent planning cycles. 
% In doing so, the agent is equipped with the all history information with lea, thereby enhancing decision-making efficacy in subsequent turns. 
In this way, the agent can access all historical information rather than just focusing on the most recent turns. Moreover, these historical details occupy only a small number of tokens, this reduces inference cost, speeds up inference, and improves decision efficiency in subsequent turns. 
The framework of D-PoT is shown in Figure~\ref{fig:framework}.

\subsection{Planning Initialization}\label{sec:plan_init}
In pursuit of the task goal $g$, the LLMs engage in $k$ turns of interactions until task completion. 
Specifically, at each turn $i$ ($i=1,\ldots, k$), the LLMs $f$ processes the visual input $x^{(i)}_{v}$ (i.e., the current screenshot) and the textual input $x^{(i)}_{t}$. It then generates the plan $p_i$ and identifies the optimal step $s^i \in p^i$ to execute:
\begin{equation}
(p^{(i)}, s^{(i)})  = f(x^{(i)}_{v}, x^{(i)}_{t}),
\end{equation}%
where the textual input $x^{(i)}_{t}$ consists of the task goal $g$, screen caption $x^{(i)}_{c}$, and execution history $x^{(i)}_{h}$. 

The textual input is further wrapped with prompts (Appendix \ref{sec:dp_pormpt}) before feeding the LLMs along with the visual input.
Concretely, we articulate our task goal at the text's outset by prompting ``\textit{Your ultimate goal is: <$g$>}''. 
Subsequently, we append the screen caption results under the heading ``\textit{The current on-screen input is: <$x^{(i)}_{c}$>}''. 
Then, we include execution history, structured as ``\textit{Here are previous actions: <$x^{(i)}_{h}$>}''. 

After feeding the inputs, we request the LLMs to generate a plan $p^{(i)} = [p_1^{(i)}, p_2^{(i)}, \ldots]$, which consists of a sequence of steps to achieve the ultimate goal.
Within those steps, the LLMs is also required to identify the optimal step $s^{(i)} \in p^{(i)}$. 
% The resulting output above is structured as ``\textit{plan: <$p^i$>, step: <$s^i$>}''. 

% output the plan $p_i$ in a JSON format, encompassing the planning and selection steps for execution. The specified output format is denoted as ``\textit{\{`plan': `...<Your Plan Here>', `step': `...<Your Step Here>'\}}''. 

% Previous research underscores the significance of interaction history for the agent \citep{zhang2023look}. Given that multiple steps occur on the similar screen within an episode, the agent is directed to prioritize interaction history outlined in the system prompt.

% The planning initialization of the DP-Agent is as follows. At the step i, the user provide an image $i_i\in \mathcal{I}$. We use prompt template $f$ to convert image $i_i$, instruction $g \in \mathcal{L}$, interaction history $h_i \in \mathcal{L}$, and more into text prompt $u_i \in \mathcal{L}$, where $\mathcal{I}$ is the mobile screen space and $\mathcal{L}$ is the language space. Following these inputs, the agent generates a plan $p_i \in \mathcal{L}$ and a selected step $s_i \in p_i$. The plan and step follow the json format.
% \begin{equation}
%     u_i = f(i_i, g_i, h_i)
% \end{equation}
% \begin{equation}
%  JSON(p_i, s_i) = Agent(i_i, u_i)
% \end{equation}

\begin{table}[ht]
\small
    \centering
    \setlength{\tabcolsep}{3mm}
    \begin{tabular}{lc}
    \toprule
        \textbf{Action Type} & \textbf{Action Description} \\ \midrule
        Click & Idx \\
        Scroll & Direction (up, down, left and right) \\
        Typ & Text \\
        Navigate & Home / Back \\
        Status & Complete \\
        Press & Enter \\
        \bottomrule
    \end{tabular}
    \caption{Six types of available actions.}
    \label{tab:available_actions}
    \vspace{-3mm}
\end{table}

\begin{figure}
    \centering
    \includegraphics[scale=0.17]{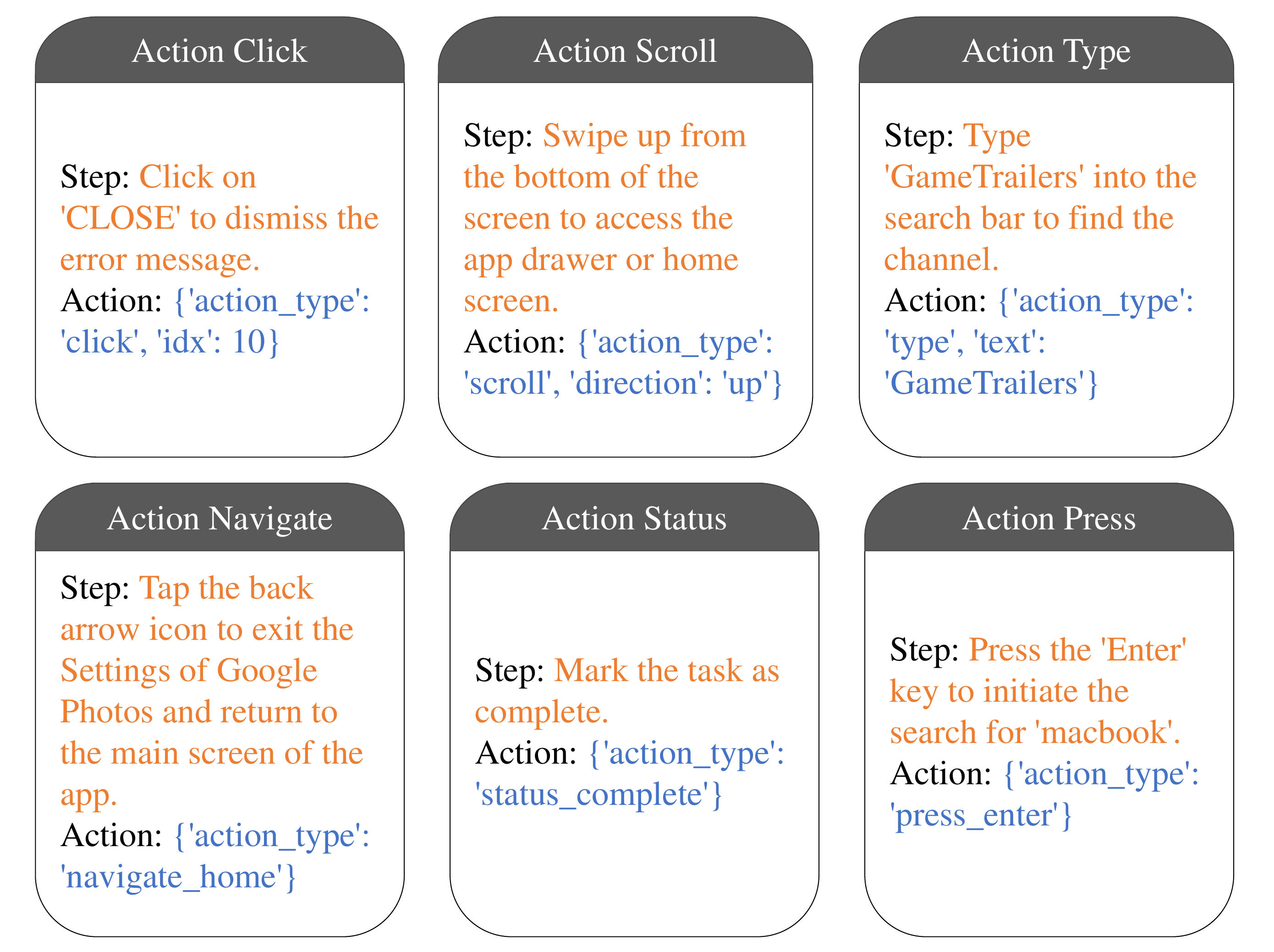}
    \caption{Examples of six types of available actions.}
    \label{fig:exam_action}
    \vspace{-5mm}
\end{figure}

In practice, $s^{(i)}$ is confined to a finite set of available actions in the GUI automation task and will be transformed into the JSON format for execution. Following \citet{rawles2023android}, we utilize six distinct types of actions as presented in Table~\ref{tab:available_actions}. There is no overlap between the different actions. Examples are provided in Figure~\ref{fig:exam_action}.

% Regarding the prompt for action, a set of 6 available actions $\mathcal{A}$ is provided, as outlined in Table~\ref{tab:available_actions}. The agent is then prompted to select the action $a^i \in \mathcal{A}$ based on the output format of the available actions.

% \subsection{Planning Execution}

% Upon completion of Planning Initialization, the subsequent task involves extracting the step $s^i$ from the return result of the initial stage. Following this, we will add the step $s^i$ to the textual input $x^{(i)}_{t}$ used during the Planning Initialization stage, following the template ``\textit{Plan: $s^i$}''. Then we get the new textual input $x^{(i)}_{t'}$.

% Regarding the prompt for action, a set of 6 available actions $\mathcal{A}$ is provided, as outlined in Table~\ref{tab:available_actions}. The agent is then prompted to select the action $a^i \in \mathcal{A}$ based on the output format of the available actions.

% \begin{equation}
% a^i  = f(x^{(i)}_{v}, x^{(i)}_{t'}| \mathcal{A}),
% \end{equation}%

% The planning execution of the DP-Agent is as follows. At the step i, we choose the step $s_i$ provided by planning initialization. We add the step to the text prompt $u_i$. Following the new instruction, the agent generate an action $a\in \mathcal{A}$, where $\mathcal{A}$ is the available actions space. We denote the process as a record:
% \begin{equation}
%  a_i = Agent((i_i, u_i, s_i)|\mathcal{A})
% \end{equation}

\subsection{Dynamic Planning Adjustment}

After the execution of $s^{(i)}$, the LLMs becomes anchored in the subsequent interaction turn with an updated visual input $x^{(i+1)}_{v}$ (e.g., a new screenshot). Simultaneously, we refine the execution history $x^{(i+1)}_h$ by concatenating $x_h^{(i)}$ and $s^{(i)}$: 
\begin{equation}
 x_h^{(i+1)} = \textsc{Concat}(x_h^{(i)}, s^{(i)}),
\end{equation}
where \textsc{Concat} denotes the concatenation operation between strings.

Consequently, the execution history is organized with consecutive elements in the format of ``\textit{step <turn id>: <action>}''. This updated execution history $x_h^{(i+1)}$ is subsequently employed according to the planning initialization process outlined in Section \ref{sec:plan_init} for turn $(i+1)$ until the task reaches completion. The task is considered complete when $i=k$ or the LLMs predicts the ``Status'' action type with the ``Complete'' action description.

\section{Experiments}

\subsection{Dataset and Setup}\label{sec:dataset}
We utilize the popular AITW dataset~\citep{rawles2023android} for evaluating our D-PoT. More details about the AITW dataset are in Appendix~\ref{sec:AITW_statistics}.
We sampled 60 episodes from each subset for analysis to get more convincing results, and incorporated screen caption results into textual input, detecting GUI icons using OCR and IconNet~\citep{sunkara2022better}. Each GUI icon is associated with a bounding box and OCR-detected text.
%\subsection{Metrics}
In line with prior research~\citep{zhang2023look, yan2023gpt4v}, our primary evaluation metric is the screen-wise action matching score, computed as the ratio of correct actions to the episode length. More details are shown in Appendix~\ref{sec:evaluation}.

\subsection{Baseline}
\label{sec:baseline}

To verify the proposed D-PoT, we used several recent agent methods as our comparison systems:
%~\citep{rawles2023android, zhang2023look}.

$\bullet$ \textbf{PaLM-2 ZS}~\citep{rawles2023android}: This setting evaluates the zero-shot performance of PaLM-2 by providing a textual description of the screen and prompting it to predict an action from the supported actions in AITW.

$\bullet$ \textbf{ChatGPT 5-shot}~\citep{zhang2023look}: ChatGPT's performance is assessed with a 5-shot prompt format similar to PaLM-2. The experiments are conducted using the ChatGPT API.

% $\bullet$ \textbf{Fine-tuned Llama-2}~\citep{zhang2023look}: The LLaMa-2 is fine-tuned with LoRA, utilizing user instructions and screen descriptions in HTML syntax, which aligns with the format used for in-context learning. The model is fine-tuned using 1\% randomly sampled training data to facilitate adaptation to the task.

$\bullet$ \textbf{GPT-4V ZS}: Zero-shot prompting with GPT-4V. The model is presented with a screenshot image and a textual description of the screen, tasked with predicting an action from the available actions.

$\bullet$ \textbf{GPT-4V 4FS}: Few-shot prompting with 4 examples. The model is presented with a screenshot image and a textual description of the screen, tasked with predicting an action from the available actions.

$\bullet$ \textbf{GPT-4V ReAct}: It represents that the interaction method of LLM is ReAct, which includes a history input of 4 turns. The inputs of every turns are screenshots, goals, and screen captions.

$\bullet$ \textbf{GPT-4V Reflexion}~\citep{shinn2024reflexion}: It represents that the interaction method of LLM is Reflexion. The model is presented with a screenshot image and a textual description of the screen, tasked with predicting an action from the available actions. When the executed action is incorrect, the action will be re-predicted.

$\bullet$ \textbf{SeeAct}~\citep{zheng2024gpt}: It represents that the interaction method of LLM is SeeAct. We choose the Text Choice method for SeeAct. The model is presented with a screenshot image and a textual description of the screen, tasked with predicting an action from the available actions.

\subsection{Implementation Details}\label{sec:implementation_detail}
We use the GPT-4V~\citep{achiam2023gpt} interface provided by OpenAI as the backbone of our agent. The GPT-4V model we use is ``gpt-4-vision-preview''. We set the ``max\_tokens'' as 300 and the ``temperature'' as 0.
We also fine-tune public large models, i.e., Llama2-7B~\citep{touvron2023llama} and LLaVa-7B~\citep{liu2023llava}, to verify the general effectiveness of our approach. For the finetuning experimental setup, training epochs are set as 3, without eval set between epochs. The maximum length of the input sequence is 2560 tokens. Text input includes the goal, screen descriptions in HTML syntax, and execution history. For inputs with a ``Plan'' experimental group, the step is spliced at the end of the input. The fine-tuning results of these open source LLMs we put in Section~\ref{sec:adaptation}.

\subsection{Main Results}

\begin{table*}[ht]
\small
\centering
\setlength{\tabcolsep}{1.95mm}
% \resizebox{\textwidth}{!}
{
\begin{tabular}{l|c|ccccc}
    \toprule
 \textbf{Model} & \textbf{Overall} & \textbf{General} & \textbf{GoogleApps} & \textbf{Install} & \textbf{Single} & \textbf{WebShopping} \\
    \midrule
    % Fine-tuned Llama 2~\citep{zhang2023look} & 28.40 & 28.56 & 30.99 & 35.18 & 27.35 & 19.92 \\ \midrule
    PaLM-2 ZS~\citep{rawles2023android} & 30.9 & - & - & - & - & - \\
    ChatGPT 5-shot~\citep{zhang2023look} & 7.72 & 5.93 & 10.47 & 4.38 & 9.39 & 8.42 \\ \midrule
    GPT-4V ZS & 34.66 & 29.69 & 35.75 & 43.50 & 32.95 & 31.42 \\
    GPT-4V 4FS & 39.71 & 34.90 & 34.97 & \textbf{50.10} & 41.62 & 36.96 \\
    GPT-4V ReAct & 42.73 & 36.20 & 42.49 & 46.60 & 49.13 & \textbf{39.22} \\
    GPT-4V Reflexion & 41.96 & 32.03 & 47.67 & 44.66 & 46.43 & 39.01 \\
    GPT-4V SeeAct & 39.58 & 34.11 & 39.38 & 40.00 & 46.24 & 38.19 \\
    D-PoT & 46.47 & 40.10 & \textbf{49.74} & 47.18 & \textbf{58.96} & 36.34 \\
    D-PoT w/ reference & \textbf{47.36} & \textbf{42.19} & 49.48 & 49.61 & \textbf{58.96} & 36.55 \\
    \bottomrule
 \end{tabular}
 }
 \caption{Main results ($\%$). Segment 1: fine-tuned Llama 2 baseline; Segment 2: in-context learning LLM baselines, ``ZS'' stands for ``zero-shot'' and ``5-shot'' stands for using 5-shot in-content learning (Section \ref{sec:baseline}); Segment 3: GPT-4V as agent model, ``D-PoT'' represents our proposed framework. ``D-PoT w/ reference'' represents seeking similar task goals during the planning initialization stage as a reference (Detailed discussion provided in Section \ref{sec:reference}). The best result is reported in boldface.}
 \label{tab:main-experiment}
 \vspace{-3mm}
\end{table*}
Table~\ref{tab:main-experiment} presents the main results of the test sets for AITW.
Based on the results, we have the following findings:
  
(i) The proposed D-PoT achieves substantial performance gains on all comparison methods in terms of Overall scores.
Particularly, D-PoT exhibits +12.7\% (34.66\% $\rightarrow$ 47.36\%) improvement on the strong baseline GPT-4V ZS. 
This presents the effectiveness of our D-PoT, that is, both environmental feedback and action history are beneficial for the GUI task. % We observe that even the GPT-4V ZS Baseline exhibits improvements over plain text input LLMs, with an increase from 7.72\%$\rightarrow$35.58\% compared to ChatGPT 5-shot and from 30.9\%$\rightarrow$35.58\% compared to PaLM-2 ZS.

% Optimizing GPT-4V agent Performance with Planning.} DP-Agent demonstrates significant performance enhancements, achieving an impressive overall improvement of +11.81\% (34.66\%$\rightarrow$46.47\%) compared to the widely adopted GPT-4V baseline.

% (ii) \textbf{Perception with the visual modality benefits the task performance.}
% We observe that even the GPT-4V ZS Baseline exhibits improvements over plain text input LLMs, with an increase from 7.72\%$\rightarrow$35.58\% compared to ChatGPT 5-shot and from 30.9\%$\rightarrow$35.58\% compared to PaLM-2 ZS.

% (i) \textbf{Enhancing Task Accuracy Through Added Image Modality.} \quad Even the GPT-4V ZS Baseline exhibits improvements over plain text input LLMs, with an increase from 7.72\%$\rightarrow$34.66\% compared to ChatGPT 5-shot and from 30.9\%$\rightarrow$34.66\% compared to PaLM-2 ZS.

% (iii) \textbf{There are 
% divergent effects of planning across specific tasks.}

(ii) We observe that our D-PoT gains improvement on the comparison methods (PaLM-2 ZS, ChatGPT 5-shot, Fine-tuned Llama-2, GPT-4V ZS, GPT-4V 4FS and GPT-4V ReAct) in almost all five categories (General, GoogleApps, Install, Single, and WebShopping). 
This indicates that our D-PoT is generalized to different GUI tasks.
%While improvements resulting from planning are steady and noticeable across the four subsets of the AITW dataset, there is a slight decrease in dynamic programming performance within the Install dataset.
% There are 
% divergent effects of planning across specific tasks.
% We notice that the impact of introducing dymanic planning varies across different tasks. In General, GoogleApps, Single, and Webshopping dataset tasks, the improvement caused by planning is stable and evident. In the Install dataset, there is little improvement in dynamic planning performance.

(iii) We observed that improvement of D-PoT on certain tasks, such as the Install and WebShopping datasets, is not significant. We think that this slight improvement may be attributed to the generated low-quality plans. 
To verify it, we select 20 episodes from the Install dataset and label them with corresponding plans (e.g., Click, Scroll, Typ, Navigate Home, Navigate Back, Press, and Complete, see Table~\ref{tab:available_actions}).
These human-annotated plans are input into LLMs instead of plans generated by GPT-4V and are prompted to select steps and predict actions.

\begin{table}[h]
\small
\setlength{\tabcolsep}{3.0mm}
    \centering
    \begin{tabular}{l c c}
    \toprule
        \textbf{Accuracy}  & \textbf{w/ GPT-4V} & \textbf{w/ Human} \\ \midrule
         Click & 17.83 & 27.39 \\
         Scroll & 0.00 & 1.27 \\
         Typ & 2.55 & 9.55 \\
         Navigate Home & 0.64 & 3.82 \\
         Navigate Back & 0.00 & 0.00 \\
         Press & 0.00 & 2.55 \\
         Complete & 2.55 & 7.64 \\ \midrule
         Total & 23.57  & \textbf{52.23} \\ 
         \bottomrule
    \end{tabular}
    \caption{Comparison of GPT-4V generated planning and human-annotated planning in the Install dataset (\%). The best average result is reported in boldface.}
    \vspace{-2mm}
    \label{tab:annotation_result}
\end{table}
Table~\ref{tab:annotation_result} shows a significant improvement for these 20 episodes, with the Total accuracy scores increasing from 23.57\%$\rightarrow$52.23\%. 
The high-quality plans are beneficial for the GUI task, which means that one of the slight decrease reasons is attributable to low-quality planning generated by GPT-4V, likely failing to stimulate this ability to generate high-quality plans during supervision fine-tuning. 

\subsection{Alleviating Planning Hallucinations and Errors}\label{sec:reference}
To mitigate planning hallucinations and errors, we additionally seek similar task goals during the planning initialization stage as a reference.
Initially, we encode the goal of each episode using sentence-transformer and identify the goals of the two most similar episodes from the remaining testsets~\cite{reimers-2019-sentence-bert}. 
We then combine the predicted actions of these two episodes as a reference for the plan. 
Additionally, we utilize InstructBlip to extract captions from the initial screen of each episode task, indicating starting point of the task~\cite{li-etal-2023-lavis}. These inputs are incorporated into the prompt for planning initialization, as outlined in the Appendix \ref{sec:dp_pormpt}.

The experimental results are shown in Table~\ref{tab:main-experiment}.
We observe that when all predicted actions from similar tasks are as a reference, the proposed D-PoT with reference gains the improvement of 0.89$\%$ accuracies on the D-PoT in terms of Overall scores. Specifically, on the General and Install datasets, incorporating references result in accuracy improvements of 2.09$\%$ and 2.43$\%$, respectively. 
This indicates that \textbf{D-PoT is effective at alleviating planning hallucinations and errors.}

\begin{table*}[t]
\centering
\small
\setlength{\tabcolsep}{1.82mm}
{
\begin{tabular}{l|ccc|c|ccccc}
    \toprule
    \textbf{Methods} & \textbf{\makecell{Static\\Planning}} & \textbf{\makecell{Dynamic\\Planning}} & \textbf{\makecell{Updating\\History}} & \textbf{Overall} & \textbf{General} & \textbf{GoogleApps} & \textbf{Install} & \textbf{Single} & \textbf{WebShopping} \\
    \midrule
    NP & \ngmark & \ngmark & \ngmark & 32.45 & 28.03 & 40.32 & 33.79 & 26.98 & 33.13 \\
    ReAct & \ngmark & \ngmark & \okmark & 38.97 & 35.61 & 46.77 & 37.24 & 39.68 & \textbf{35.54} \\ 
    SP & \okmark & \ngmark& \ngmark & 28.58 & 21.97 & 38.71 & 20.69 & 42.86 & 18.67 \\
    DP & \ngmark & \okmark& \ngmark & 42.01 & 31.06 & 50.81 & 40.69 & \textbf{57.14} & 30.72 \\
    D-PoT & \ngmark & \okmark & \okmark & \textbf{45.75} & \textbf{45.45} & \textbf{52.42} & \textbf{44.14} & 52.38 & 34.34 \\
    \bottomrule
 \end{tabular}
 }
 \caption{The ablation studies on planning mechanisms. Static Planning: Create a plan based on the provided screenshot and goal at the outset of the episode. And utilize this plan consistently throughout the episode to direct LLMs in predicting actions; Dynamic Planning: Continuously adapt and formulate plans during task execution, considering all available input information; Updating History: Incorporate the steps into the execution history and utilize them in the planning process. Each experiment's execution or omission of a particular process is denoted by $\okmark$ (if performed) or $\ngmark$ (if not performed). The best average result is in boldface.}
 \label{tab:ana-experiment}
\end{table*}
\subsection{Ablation Study of Varied Planning}
%{Comparison between No Planning, Static Planning and Dynamic Planning}\label{sec:planning}

To study the impact of dynamic planning, we randomly sampled 20 episodes from each data subset, with a total of 100 episodes as the dataset for the ablation experiment, and built several baselines.

\noindent $\bullet$ \textbf{No Planning (NP)}: Its inputs are screenshots, goals, and screen captions. We ask the LLMs to predict actions directly based on these inputs without specifying a plan.

\noindent $\bullet$ \textbf{ReAct}: It represents that the interaction method of LLM is ReAct, which includes a history input of 4 turns. The inputs of every turns are screenshots, goals, and screen captions.

\noindent $\bullet$ \textbf{Static Planning (SP)}: It represents the utilization of planning statically. We will ask LLMs to generate a plan at the beginning of the episode and add the plan to the prompt during the whole episode.

\noindent $\bullet$ \textbf{Dynamic Planning (DP)}: It represents the utilization of planning, excluding selecting steps and updating execution history. The inputs of DP are screenshots, goals, and screen captions. When receiving a new screenshot, we ask LLMs to generate a plan and then take action.

Table~\ref{tab:ana-experiment} presents the detailed results of the test set for the AITW dataset. 
First, the accuracy scores of DP and D-PoT are higher than those of NP, SP and ReAct. This means that dynamic planning is significantly superior to static planning in the graphical user interface automation task. We think that this superiority contributes to two potential or possible factors: 1) This planning greatly stimulates the understanding ability of the LLMs-based agent for the graphical user interface automation task; 2) Throughout task execution, the historical information extracted by steps helps the LLMs-based agent flexibly update its plan for the environment changes and unseen scenarios, especially compared to ReAct, reduces inference cost and greatly improves the performance.

Second, the comparison among NP, ReAct, DP, and D-PoT reveals that integrating planning leads to substantial enhancements preceding the predicted action. We think that this effect arises as the generative planning may prompt LLMs to engage in GUI automation, thereby enhancing their comprehension of the intended goal. This demonstrates that the proposed D-PoT obtains notable enhancement via plan integration before action prediction.

Third, we observe that D-PoT outperformed DP in terms of Overall scores. 
This indicates that incorporating execution history into LLMs enhances GUI automation through dynamic planning.
In other words, historical information is beneficial for LLMs in GUI automation, especially dynamic planning based on historical steps. 
% Moreover, the accuracy scores of D-PoT are inferior to those of DP on the Single and Install datasets. 
Moreover, the accuracy scores of D-PoT are inferior to those of DP on the Single datasets. 
In addition to the generated low-quality plans in Table~\ref{tab:annotation_result}, part of the reasons may be that the short episode length reduces the reliance on historical information for the Single dataset.
% In addition to the generated low-quality plans in Table~\ref{tab:annotation_result}, part of the reasons may be that the short episode length reduces the reliance on historical information for the Single dataset, and D-PoT may be misled by incorrect historical information for the Install dataset.

\subsection{Exploring the Proportion and Correct Rate of Predicted Actions}\label{sec:main_exploring}

To conduct a detailed analysis of the impact of dynamic planning, we dive into the correct rate and the proportion of predicted actions. 
Specifically, we combine five categories for an overarching analysis, more details of the correct rate of predicted actions are in Appendix~\ref{sec:ablation_appendix}. 
% and compute the proportion of actions within the dataset in Table~\ref{tab:analyse_statis}. 
Table~\ref{tab:analyse_correct} presents the overall predicted ratio and the predicted accuracy ratio for different actions. Due to the potential occurrence of unpredictable actions in LLMs, it's possible that the sum of predicted probabilities may not equal 1.
% \begin{table}[ht]
% \small
% \setlength{\tabcolsep}{6.8mm}
%     \centering
%     \begin{tabular}{l | c}
%     \toprule
%         \textbf{Category}  & Proportion (\%) \\ \midrule
%          Click & 52.54 \\
%          Scroll & 13.97 \\
%          Typ & 10.67 \\
%          Navigate Home & 4.44 \\
%          Navigate Back & 0.79 \\
%          Press & 1.75 \\
%          Complete & 15.87 \\ \bottomrule
%     \end{tabular}
%     \caption{The proportion of actions on AITW.}
%     \vspace{-5mm}
%     \label{tab:analyse_statis}
% \end{table}

\begin{table*}[ht]
    \centering
    \small
    \setlength{\tabcolsep}{2.2mm}
    {
    \begin{tabular}{l|ccccccc}
    \toprule
          \textbf{Model} & \textbf{Click} & \textbf{Scroll} & \textbf{Typ} & \textbf{Home} & \textbf{Back} & \textbf{Press} & \textbf{Complete} \\
        \midrule
        NP & 78.57 / 26.67 & 4.29 / 1.27 & 4.29 / 2.38 & 4.13 / 1.11 & 1.59 / 0 & 0.48 / 0.16 & 1.43 / 1.43  \\
        ReAct & 70.79 / 27.14 & 4.92 / 1.59 & \textbf{7.94 / 4.92} & 3.02 / 1.43 & \textbf{1.90 / 0.16} & 2.54 / 0.32 & 3.33 / 3.02  \\
        SP & 71.11 / 19.84 & 8.41 / 1.43 & 1.43 / 0.63 & 8.73 / 0.48 & 2.06 / 0 & \textbf{2.7 / 0.48} & 3.81 / 3.33  \\
        DP & 71.90 / 25.08 & 3.49 / 1.11 & 3.17 / 2.54 & \textbf{6.98 / 2.06} & 0.63 / 0 & 1.59 / 0.32 & 11.75 / 8.57 \\
        D-PoT & 68.45 / 26.83 & \textbf{7.01 / 2.7} & 3.51 / 3.02 & 1.03 / 0.79 & 0.41 / 0 & 0.62 / 0 & \textbf{18.14 / 9.52} \\
        D-PoT w/ reference & \textbf{64.29 / 29.37} & \textbf{10.48 / 2.7} & 5.87 / 3.33 & 2.38 / 0.95 & 0.95 / 0 & 3.49 / 0.16 & 11.9 / 8.10 \\
        \bottomrule
    \end{tabular}}
    \caption{The predicted ratio and the predicted accuracy ratio for different actions($\%$). the number on the left of ``/'' is the predicted ratio, and the number on the right of ``/'' is the predicted accuracy ratio. The best result is in boldface.}
    \label{tab:analyse_correct}
    \vspace{-2mm}
\end{table*}

Our observations based on these statistics reveal the following two findings:

(i) \textbf{Dynamic planning empowers LLMs to enhance their task management capabilities.} \quad Within the DP and D-PoT experimental groups, we observed a noteworthy increase in both the prediction proportion and accuracy rate of ``Complete'' actions. 
    This suggests that dynamic planning enhances the comprehension of LLMs-based agent in the current task.

(ii) \textbf{Dynamic planning reduces the invalid predicted click action.} \quad  We observed a significant decrease in the prediction ratio for ``Click'' with the introduction of dynamic planning, but the prediction accuracy rate is not affected.
    Existing work indicates that GPT-4V is more likely to predict the ``Click'' action~\cite{yan2023gpt4v}. However, the proposed D-PoT minimizes invalid and erroneous click actions, showcasing a better comprehension of the implementation progress of the current plan.

\subsection{Adaptation to Unfamiliar Tasks}
\label{sec:adaptation}
As new applications continually emerge, their interfaces often pose unfamiliarity to agents. 
Despite the diversity of GUI tasks, there exists a semblance of similarity in screen navigation logic. 
Even when the interface is unknown, certain screen transition patterns remain consistent. Consequently, the proposed D-PoT utilizes dynamic planning to capture environmental changes and historical steps, which will be beneficial for adaptation to unfamiliar tasks.
% , aiding in the effective recall of steps that can be applied universally across different application interfaces. 
To validate this, we select two base models, Llama2-7B and LLaVa-7B, for fine-tuning. Llama2-7B serves to verify the effectiveness of the D-PoT method on plain text, while LLaVa-7B serves to verify its effectiveness on multimodal data. We randomly choose the GoogleApps dataset as the training set and the remaining datasets as the test set. The five datasets contain various task categories. We utilize both the D-PoT instruction from our method and the action instruction from AITW for fine-tuning.

\begin{table}[h]
\small
    % \resizebox{.75\columnwidth}{!}
    \setlength{\tabcolsep}{0.3mm}
    \centering
    \begin{tabular}{l c c c c}
        \toprule
        \textbf{Methods} & \textbf{General} & \textbf{Install} & \textbf{Single} & \small \textbf{WebShopping} \\ \midrule
        Llama2-7B\\
        \quad w/ all data & 28.56 & 35.18 & 27.35 & 19.92 \\ \midrule
        Llama2-7B \\
        \quad  NP Baseline & 13.08 & 17.12 & 3.87 & 8.71 \\
        %\quad w/ Plan  &  &  &  &  \\
        \quad  Plan by GPT-4V & \textbf{24.67} & \textbf{23.46} & \textbf{39.48} & \textbf{19.48} \\
        \quad  Plan by Itself & 17.81 & 17.58 & 15.87 & 12.46 \\
        \midrule
        LLaVa-7B\\
        %\quad w/ Plan & & & & \\
        \quad  NP Baseline & 17.81 & 17.98 & 1.66 & 10.91 \\
        \quad  Plan by GPT-4V & 27.19 & 26.77 & 44.46 & 20.61 \\
        \quad  Plan by Itself & \textbf{30.73} & \textbf{29.39} & \textbf{45.94} & \textbf{21.67} \\
        \bottomrule
    \end{tabular}
    \caption{Finetuning results of Llama2-7B and LLaVa-7B. Segment 1: ``w/ all data'' stands for the model is fine-tuned with 1\% randomly sampled training data to help adapt to this task~\citep{zhang2023look}. Segments 2 \& 3: The training set is 180 episodes in the GoogleApps, and the test set is 180 episodes in other datasets. ``\textit{GPT-4V}'' stands for planning is made by GPT-4V. ``\textit{itself}'' stands for planning made by the finetuned model itself. The best average result is in boldface.}
    \label{tab:finetuning_result}
    \vspace{-2mm}
\end{table}

% , including the results for Llama2-7B and LLaVa-7B which are fine-tuned using nearly all the action instruction data. 
The results in Table~\ref{tab:finetuning_result} indicate that LLMs fine-tuned with D-PoT data exhibit significant improvements in other tasks and demonstrate robust adaptability to unknown tasks compared to direct fine-tuning with action instructions. 
Even on the Llama2-7B model, the experimental results of fine-tuning using only a small amount of D-PoT data are comparable to those of fine-tuning using the full AITW dataset. 
This verifies the effectiveness of D-PoT for out-of-domain tasks.

Additionally, in the experiment with LLaVa-7B, we observed that allowing LLaVa-7B to learn dynamic planning rather than following the planned prediction actions formulated by GPT-4V, yielded higher accuracy scores. 
This indicates that our fine-tuned LLaVa-7B model learned the plan from the GoogleApp dataset and is capable of planning effectively for tasks in other domains. This further supports the notion that \textbf{D-PoT can adapt LLMs to unfamiliar tasks.}

\subsection{Error Analysis} 
% We have observe that even the strongest GPT-4V encounters numerous challenges in dynamic planning. Upon analyzing the experimental results, we attribute the influence of GPT-4V to two main factors: 1) GPT-4V exhibits inherent biases in its plan to mobile phone task datasets. 2) The instructions provided in the current dataset remain insufficient. 
% Further analysis and specific error sample analysis are provided in Appendix ~\ref{sec:error_example}.
% We believe that these analyses will help future researchers in comprehending the limitations of GPT-4V and other LLMs in mobile tasks, particularly within the AITW benchmark. Additionally, they may offer inspiration for the development of subsequent mobile task datasets.
To dive into the mistakes of GPT-4V in dynamic planning and facilitate future studies, we categorize three common errors that lead to discrepancies between the predictions of GPT-4V and human-annotated predictions. 
% Due to space constraints, we present only one of the errors in the main body. 
More details are presented in Appendix~\ref{sec:error_example}.

% We believe that these analyses will help future researchers comprehend the limitations of GPT-4V and other LLMs in mobile tasks, particularly within the AITW benchmark. Additionally, they may offer inspiration for the development of subsequent mobile task datasets.

\section{Conclusion}

This study introduces a prompting approach called D-PoT, designed to facilitate interactions in a multimodal environment. 
% We capitalized on the potent zero-shot capabilities of GPT-4V to evaluate the accuracy of LLMs in predicting execution actions on smartphones screenshots. 
D-PoT encourages LLMs to dynamically update planning based on feedback from the environment and execution history. Through the application of D-PoT, we demonstrate that the D-PoT surpasses the widely adopted GPT-4V baseline on the AITW benchmark dataset. Meanwhile, our findings indicate that the D-PoT excels in adapting to unfamiliar tasks, and can predict different actions more correctly.

\section*{Limitations}
This study utilizes the powerful zero-shot capability of LLMs to forecast smartphone actions by incorporating prompt constraints. Our focus lies predominantly on exploring the efficacy of dynamic planning in enhancing action prediction within a given scenario during an episode.
In terms of social impact, employing LLM-based agents on mobile phones holds promise for assisting individuals with disabilities. It's worth noting that applying LLMs-based agents on smartphones presents certain constraints. While we find promise in the observed improvement in predicted action accuracy over longer episodes through dynamic planning, practical implementation remains a distant goal. Many challenges stem from the limited knowledge of the mobile phone domain within LLMs, highlighting inherent imperfections. These issues warrant further investigation in future research endeavors.

\section*{Acknowledgements}
We want to thank all the anonymous reviewers for their valuable comments. This work was supported by the National Natural Science Foundation of China (62376075, 62276077, U23B2055 and 62406188),  Guangdong Basic and Applied Basic Research Foundation (2024A1515011205), Shenzhen College Stability Support Plan under Grants (GXWD20220811170358002 and GXWD20220817123150002), and the project of National key research and development program of China (No. 2023YFC3804600).
% We extend our heartfelt appreciation to the anonymous reviewers for their professional, insightful, and constructive comments. Additionally, we gratefully acknowledge the support of .

% \bibliographystyle{acm}
\bibliography{latex/custom}

\newpage
\appendix
\onecolumn

\section{Example Appendix}
\subsection{Dynamic planning prompting}\label{sec:dp_pormpt}

We use the following prompt for Planning Initialization.

\begin{lstlisting}[style=prompt]
Imagine that you are a robot operating a mobile. Like how humans operate the mobile, you can click on the screen, type some text, go home, go back to the last screen, scroll up, down, left and right, or mark the status as complete. Given a goal and a mobiel screen, you need to make a plan to achieve your goals based on the current screen, and choose the steps that should be achieved on the current screen from the plan you have made. Since achieving this goal is a **continuous process**, you will be given the **previous steps and actions** that have been performed, so please pay attention to this information. There may be multiple ways to achieve your goals, but what you need to do is create the plan that best suits your current situation based on the current screen input.

**Your ultimate goal is: check out phone information.** 
The current on-screen input is:
Screen:
<p id=0 class=``text'' alt=``vvaiipaper,''>vvaiipaper,</p>
<p id=1 class=``text'' alt=``sieep,''>sieep,</p>
<p id=2 class=``text'' alt=``iolL''>iolL</p>
<p id=3 class=``text'' alt=``SIZE''>SIZE</p>
<p id=4 class=``text'' alt=``Sound''>Sound</p>
<img id=5 class=ICON\_VOLUME\_STATE alt=``''></p>\n <p id=6 class=``text'' alt=``Volume,''>Volume,</p>
<p id=7 class=``text'' alt=``vibration,''>vibration,</p>
<p id=8 class=``text'' alt=``Do''>Do</p>
<p id=9 class=``text'' alt=``Not''>Not</p>
<p id=10 class=``text'' alt=``Disturb''>Disturb</p>\newline <p id=11 class=``text'' alt=``Storage''>Storage</p>
<p id=12 class=``text'' alt=``used''>used</p>
<p id=13 class=``text'' alt=``GB free''>GB free</p>
<p id=14 class=``text'' alt=``49\%''>49\%</p>
<p id=15 class=``text'' alt=``-32.63''>-32.63</p>
<p id=16 class=``text'' alt=``Privacy''>Privacy</p>
<p id=17 class=``text'' alt=``Permissions,''>Permissions,</p>
<p id=18 class=``text'' alt=``account''>account</p>
<p id=19 class=``text'' alt=``personal''>personal</p>
<p id=20 class=``text'' alt=``data''>data</p>
<p id=21 class=``text'' alt=``activity,''>activity,</p>
<p id=22 class=``text'' alt=``Location''>Location</p>
<img id=23 class=ICON\_LOCATION alt=``''></p>
<p id=24 class=``text'' alt=``On''>On</p>
<p id=25 class=``text'' alt=``have access''>have access</p>
<p id=26 class=``text'' alt=``- 4 apps''>- 4 apps</p>
<p id=27 class=``text'' alt=``location''>location</p>
<p id=28 class=``text'' alt=``to''>to</p>
<p id=29 class=``text'' alt=``Security''>Security</p>
<p id=30 class=``text'' alt=``lock, fingerprint''>lock, fingerprint</p>
<p id=31 class=``text'' alt=``Screen''>Screen</p>
Here are previous actions: (format: action \u2192 action description)
Previous Actions:
{''step\_idx'': 0, ''action\_description'': ''scroll up''}
{''step\_idx'': 1, ''action\_description'': ''click []''}
{''step\_idx'': 2, ''action\_description'': ''scroll up''}
And the previous steps:
Previous Steps:
Step 1. Swipe up from the bottom of the screen to access the app drawer.
Step 2. Tap on the 'Settings' icon to open the settings menu.
Step 3. Scroll up to reveal more settings options.

Please formulate an operational guide for future operations for solving the goal. The guide includes:
1. Plan: A **multi-step future** plan **(start from current screen, DON'T include previous steps)**; steps indexed by numbers.
2. Step: Based on the current screen and Previous Steps, provide the **immediate** step that needs to be taken from the Plan.
"**Output Format:** A JSON dictionary strictly following the format: "{'plan': '...<Your Plan Here>', 'step': '...<Your Step Here>'} "If the goal has already been implemented, no more planning is required, Provide {'plan': '1. Mark the task as complete', 'step': 'Mark the task as complet'}.
**Please do not output any content other than the JSON format.**    
\end{lstlisting}
We use the following prompt for Planning Initialization with references.
\begin{lstlisting}[style=prompt]
Imagine that you are a robot operating a mobile. Like how humans operate the mobile, you can click on the screen, type some text, go home, go back to the last screen, scroll up, down, left and right, or mark the status as complete. Given a goal and a mobiel screen, you need to make a plan to achieve your goals based on the current screen, and choose the steps that should be achieved on the current screen from the plan you have made. Since achieving this goal is a **continuous process**, you will be given the **previous steps and actions** that have been performed, so please pay attention to this information. There may be multiple ways to achieve your goals, but what you need to do is create the plan that best suits your current situation based on the current screen input.
**Your ultimate goal is: What is the price of a 12' ladder at Home Depot?.**
I also give you two similar examples as a reference, here are their goal, the initial caption of mobile screen, and all the execution actions to complete goal:
Goal: What's the price of the 1000-Watt EGO Power+ Snow Blower?
Caption: The information on the phone screen is a screenshot of the Google Play Store, displaying various apps available for download. The screenshot provides a visual representation of the apps that can be found on the Google Play Store, allowing users to easily browse and choose from a variety of options.
Execution history: {\"step_idx\": 0, \"action_description\": \"click [9]\"}

{\"step_idx\": 1, \"action_description\": \"click [9]\"}

{\"step_idx\": 2, \"action_description\": \"click []\"}

{\"step_idx\": 3, \"action_description\": \"type\"}

{\"step_idx\": 4, \"action_description\": \"press_enter\"}

{\"step_idx\": 5, \"action_description\": \"click [Shopping]\"}

{\"step_idx\": 6, \"action_description\": \"scroll up\"}

{\"step_idx\": 7, \"action_description\": \"click [Official Site - Shop Ego Lb5300]\"}

\\{\"step_idx\": 8, \"action_description\": \"status_complete\"\\}

Goal: What's the price of the new iPhone on eBay?
Caption: The information displayed on the phone screen is a screenshot of the Google Calendar app. The screenshot shows the current date and time, as well as a list of upcoming events for the next few days. It also highlights some of the features of the Google Calendar app, such as the ability to add events, set reminders, and manage multiple calendars. The screenshot provides an overview of the user's schedule and helps them stay organized and on top of their upcoming events.
Execution history: {\"step_idx\": 0, \"action_description\": \"click [9]\"}

{\"step_idx\": 1, \"action_description\": \"click [weather like in]\"}

{\"step_idx\": 2, \"action_description\": \"click [google.com/search?q=wea]\"}

{\"step_idx\": 3, \"action_description\": \"type\"}

{\"step_idx\": 4, \"action_description\": \"click [iPhone on]\"}

{\"step_idx\": 5, \"action_description\": \"scroll up\"}

{\"step_idx\": 6, \"action_description\": \"click [iPhones for Sale - New & Used]\"}

{\"step_idx\": 7, \"action_description\": \"scroll up\"}

{\"step_idx\": 8, \"action_description\": \"click [H]\"}

\\{\"step_idx\": 9, \"action_description\": \"status_complete\"\\}

The current on-screen input is:
Screen: <p id=0 class=\"text\" alt=\"Mon, Oct 10\">Mon, Oct 10</p>
<img id=1 class=ICON_CLOUD alt=\"\"></p>
<p id=2 class=\"text\" alt=\"56\u00b0F\">56\u00b0F</p>
<img id=3 class=ICON_CALL alt=\"\"></p>
<img id=4 class=ICON_CHAT alt=\"\"></p>
<img id=5 class=ICON_PLAY alt=\"\"></p>
<img id=6 class=ICON_GOOGLE alt=\"\"></p>
<img id=7 class=ICON_MIC alt=\"\"></p>
<img id=8 class=ICON_NAV_BAR_RECT alt=\"\"></p>
<img id=9 class=ICON_NAV_BAR_CIRCLE alt=\"\"></p>
<img id=10 class=ICON_V_BACKWARD alt=\"\"></p>

Here are previous actions: (format: action \u2192 action description)
Previous Actions:
{'action_type': 'click', 'idx': 15}
And the previous steps:
Previous Steps:
Step 1. Press the home button to exit the email setup screen.

Please formulate an operational guide for future operations for solving the goal. The guide includes:
1. Plan: A **multi-step future** plan **(start from current screen, DON'T include previous steps)**; steps indexed by numbers.
2. Step: Based on the current screen and Previous Steps, provide the **immediate** step that needs to be taken from the Plan.
"**Output Format:** A JSON dictionary strictly following the format: "{'plan': '...<Your Plan Here>', 'step': '...<Your Step Here>'} "If the goal has already been implemented, no more planning is required, Provide {'plan': '1. Mark the task as complete', 'step': 'Mark the task as complet'}.
**Please do not output any content other than the JSON format.**    
\end{lstlisting}

\newpage
\subsection{The statistics for AITW dataset}\label{sec:AITW_statistics}
AITW is a comprehensive benchmark tailored for GUI control, comprising natural language instructions, screenshots, and associated actions. 
Agent predicts execution actions based on screenshots and task goals across five categories shown in Table~\ref{tab:dataset}. 
The dataset spans over 350 applications and websites, totaling 715,000 episodes with 30,000 unique instructions.
Subsequently, each filtered subset is partitioned episode-wise into training, validation, and test sets following 80/10/10 splits. 
\begin{table}[ht]
\small
    \centering
    \setlength{\tabcolsep}{2mm}
    \begin{tabular}{lccc}
    \toprule
        \textbf{Dataset} & \textbf{Episodes} & \textbf{Screens} & \textbf{Instructions} \\\midrule
        General & 9,476 & 85,413 & 545  \\
        Install & 25,760 & 250,058 & 688 \\
        GoogleApps & 625,542 & 4,903,601 & 306 \\
        Single & 26,303 & 85,668 & 15,366 \\
        WebShopping & 28,061 & 365,253 & 13,473 \\\bottomrule
    \end{tabular}
    \caption{Statistics for AITW dataset.}
    \label{tab:dataset}
    %\vspace{-3.6mm}
\end{table}
\subsection{Evaluation metrics}\label{sec:evaluation}
Specifically, for click actions, correctness is determined if the selected element is within a 14$\%$ screen distance from the gold gestures or falls within the same detected bounding box as the user's gestures. Given the error in OCR identification, we select the top left, top right, bottom left, bottom right, and center of the box as sample points for calculating coordinate distances. Regarding scroll actions, correctness is assessed if the selected direction aligns with the scroll direction of the user's gestures. For other actions, correctness is established only if the types of actions match~\citep{rawles2023android}.
\subsection{The correct rate of predicted actions in ablation studies}
\label{sec:ablation_appendix}

We compute the proportion of actions within the dataset in Table~\ref{tab:analyse_statis}. 

\begin{table}[ht]
\small
\setlength{\tabcolsep}{6.8mm}
    \centering
    \begin{tabular}{l | c}
    \toprule
        \textbf{Category}  & Proportion (\%) \\ \midrule
         Click & 52.54 \\
         Scroll & 13.97 \\
         Typ & 10.67 \\
         Navigate Home & 4.44 \\
         Navigate Back & 0.79 \\
         Press & 1.75 \\
         Complete & 15.87 \\ \bottomrule
    \end{tabular}
    \caption{The proportion of actions on AITW.}
    \vspace{-5mm}
    \label{tab:analyse_statis}
\end{table}

We provide the predicted action accuracy for all datasets of ablation experiments in Table~\ref{tab:total_analyse_correct}.
\begin{table*}[ht]
% \fontsize{4pt}{4pt}\selectfont
%     \centering
%     \resizebox{\textwidth}{!}
 \setlength{\tabcolsep}{4.2mm}
    \centering\small
    {
    \begin{tabular}{l l |ccccc}
    \toprule
        \textbf{Model} & \textbf{Action} & \textbf{General} & \textbf{GoogleApps} & \textbf{Install} & \textbf{Single} & \textbf{Webshopping} \\ \midrule
        \multirow{7}{*}{NP}  & Click &  23.48 & 33.06 & 25.52 & 23.81 & 26.51 \\
                            & Scroll & 0.67 & 2.42 & 2.07 & - & 0.60 \\
                            & Typ &  3.03 & - & 2.76 & - & 4.22 \\
                            & Navigate Home & - & 1.61 & 2.07 & - & 1.20 \\
                            & Navigate Back & - & - & - & - & - \\
                            & Press Enter & - & - & - & 1.59 & - \\
                            & Complete & 0.76 & 3.23 & 1.38 & 1.59 & 0.60 \\ \midrule

        \multirow{7}{*}{SP}  & Click &  16.67 & 30.65 & 14.48 & 25.40 & 16.87 \\
                            & Scroll & 3.03 & 0.81 & 2.76 & - & - \\
                            & Typ &  - & - & 0.69 & 3.17 & 0.60 \\
                            & Navigate Home & - & 0.81 & 0.69 & - & 0.60 \\
                            & Navigate Back & - & - & - & - & - \\
                            & Press Enter & - & - & - & - & - \\
                            & Complete & 2.27 & 6.45 & 2.07 & \textbf{26.98} & 0.60 \\ \midrule
                            
        \multirow{7}{*}{DP}  & Click &  16.67 & 36.29 & 24.14 & \textbf{30.16} & 22.29 \\
                            & Scroll & 0.76 & 1.61 & 2.07 & - & 0.60 \\
                            & Typ &  2.27 & 0.81 & \textbf{4.14} & \textbf{1.59} & 3.01 \\
                            & Navigate Home & \textbf{4.52} & \textbf{2.42} & \textbf{3.45} & - & \textbf{1.81} \\
                            & Navigate Back & - & - & - & - & - \\
                            & Press Enter & - & - & - & - & - \\
                            & Complete & 9.85 & 9.68 & \textbf{6.90} & 22.22 & \textbf{3.01} \\ \midrule
        % \multirow{5}{*}{PS}  & Click &  26.52 & 34.68 & \textbf{25.52} & 26.98 & 25.90 \\
        %                     & Scroll & \textbf{3.79} & 5.65 & \textbf{6.90} & 0.00 & 0.60 \\
        %                     & Type &  0.76 & 0.00 & \textbf{4.14} & \textbf{1.59} & 3.01 \\
        %                     & Navigate Home & 0.00 & 1.61 & 0.69 & 0.00 & 1.20 \\
        %                     & Complete & 6.82 & 8.87 & 4.83 & 19.05 & 2.41 \\ \midrule
        \multirow{7}{*}{D-PoT}  & Click &  \textbf{27.27} & 35.48 & 21.38 & 23.81 & 25.90 \\
                            & Scroll & 3.03 & \textbf{3.23} & \textbf{5.52} & - & 0.60 \\
                            & Typ &  \textbf{3.79} & 0.81 & 3.45 & \textbf{1.59} & \textbf{4.22} \\
                            & Navigate Home & - & 1.61 & 1.38 & - & 0.60 \\
                            & Navigate Back & - & - & - & - & - \\
                            & Press Enter & - & - & - & - & - \\
                            & Complete & \textbf{11.36} & \textbf{11.29} & 6.21 & \textbf{26.98} & \textbf{3.01} \\ \midrule
        \multirow{7}{*}{\makecell{D-PoT \\ w/ reference}}  & Click &  21.21 & \textbf{37.90} & \textbf{28.28} & 28.57 & \textbf{30.72} \\
                            & Scroll & \textbf{4.55} & \textbf{3.23} & 4.83 & - & - \\
                            & Typ &  2.77 & - & \textbf{4.14} & \textbf{6.35} & \textbf{4.82} \\
                            & Navigate Home & 0.76 & 1.61 & 1.38 & - & 0.60 \\
                            & Navigate Back & - & - & - & - & - \\
                            & Press Enter & - & - & - & - & - \\
                            & Complete & 9.85 & \textbf{11.29} & 5.52 & 22.22 & 1.2 \\ \bottomrule
    \end{tabular}
    }
    \caption{The correct rate of predicted actions of GPT-4V and D-PoT in ablation studies. We mainly collected the correct rate of ``Click'', ``Scroll'', ``Typ', ``Navigate'' and ``Complete'' actions. To make it look nice, we'll replace 0 with ``-''. The best average result is in boldface.}
    \label{tab:total_analyse_correct}
\end{table*}
\newpage
\section{Errors Examples}
\label{sec:error_example}
The three errors are shown here.

\begin{figure}[h]
    \centering
    \includegraphics[scale=0.45]{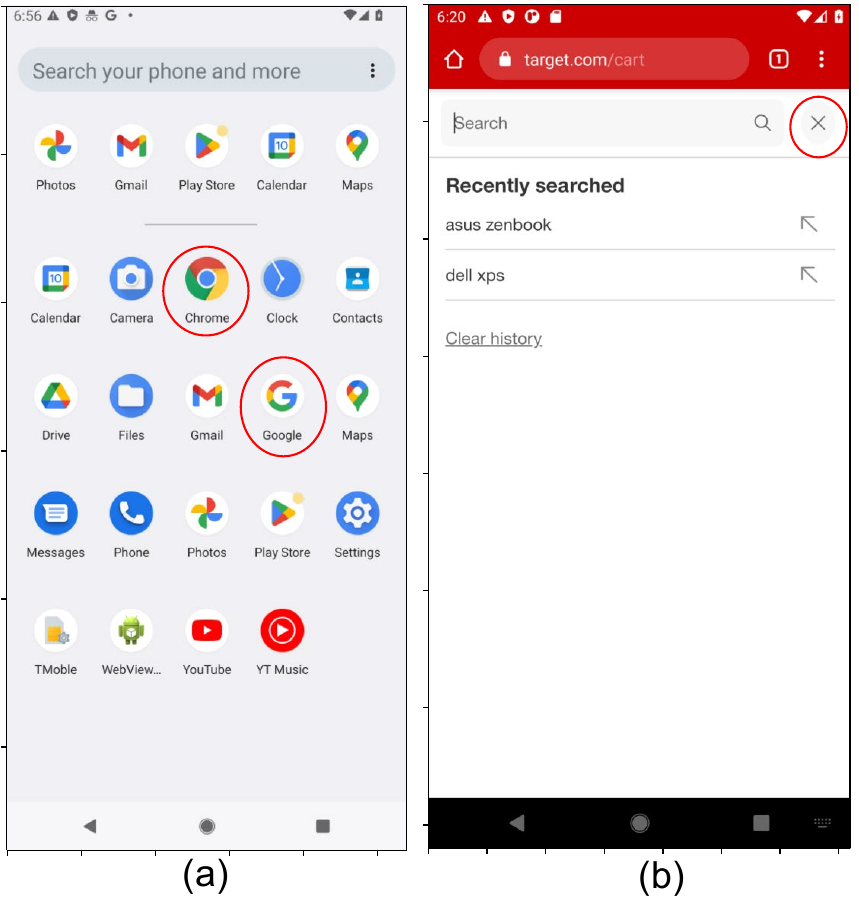}
    % \vspace{-8mm}
    \caption{The first common error is a bias of GPT-4V on mobile tasks. The red circles are the steps that GPT-4V performs in a dynamic schedule.}
    \label{fig:error1}
    \vspace{-3mm}
\end{figure}

The first common error we identify is a bias of GPT-4V on mobile tasks. GPT-4V often exhibits ``preferences'' in its planning. As illustrated in Figure~\ref{fig:error1}(a), when tasked with searching for specific information, GPT-4V tends to click on Google, while the human-annotated prediction suggests clicking on Chrome. Similarly, in Figure~\ref{fig:error1}(b), when required to input text in the search bar, GPT-4V may plan to clear the search bar first, whereas the human prediction is to directly input the text.

\begin{figure}[h]
    \centering
    \includegraphics[scale=0.45]{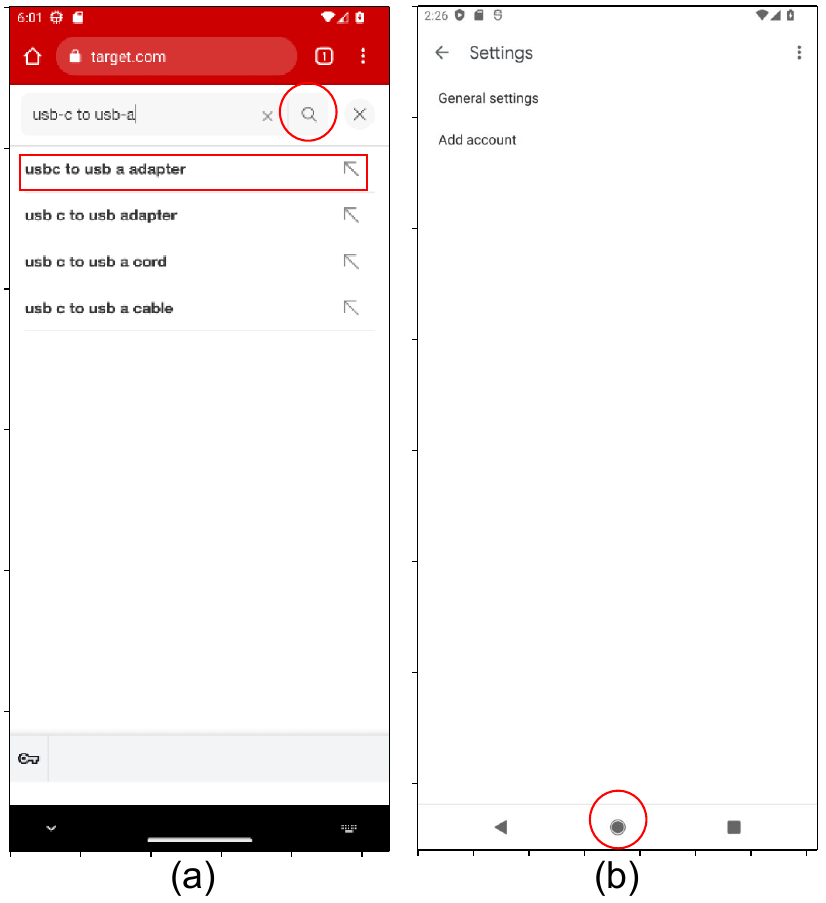}
    % \vspace{-8mm}
    \caption{The second common error we recognize is instruction overlap in the AITW dataset. The red circles are the steps that GPT-4V performs in a dynamic schedule}
    \label{fig:error2}
    \vspace{-3mm}
\end{figure}

The second common error we recognize is instruction overlap in the AITW dataset. The same operation on one mobile screen can correspond to two different actions. For instance, in Figure~\ref{fig:error2}(a), when searching for an item, GPT-4V may click on 'search' or the search entry, whereas the human prediction is to press. In Figure~\ref{fig:error2}(b), when returning to the home page, GPT-4V often clicks on the ``home'' button below, while the human instruction is to ``navigate home''.
\begin{figure}[h]
    \centering
    \includegraphics[scale=0.45]{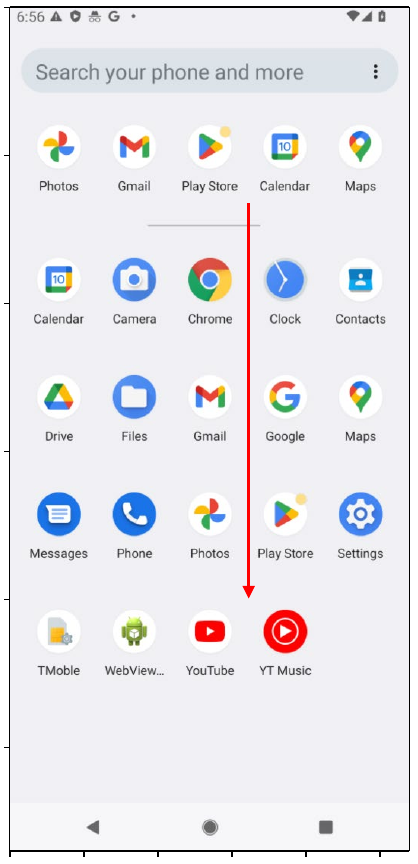}
    % \vspace{-8mm}
    \caption{The third common error we classify as confusion in gesture operations. The red arrow indicates that the GPT-4V wants to slide under in dynamic planning}
    \label{fig:error3}
    \vspace{-3mm}
\end{figure}
The third common error we classify as confusion in gesture operations. For example, in Figure~\ref{fig:error3}, when swiping down to view more apps, the corresponding gesture should be from bottom to top, indicating ``scroll up''. However, GPT-4V also suggests swiping down, but its predicted instruction is ``scroll down''.

\end{document}